# Binary Classification of Alzheimer's Disease using sMRI Imaging modality and Deep Learning

Ahsan Bin Tufail [1,2], Yong-Kui Ma [1], Qiu-Na Zhang [1]

**Correspondence Author: Yong-Kui Ma**
yk_ma@hit.edu.cn

[1] School of Electronics and Information Engineering, Harbin Institute of Technology, 92 Xidazhi St, Nangang Qu, Harbin, Heilongjiang Province, China, 150001

[2] Department of Electrical and Computer Engineering, COMSATS University Islamabad, Sahiwal Campus, COMSATS Road off G.T Road, Sahiwal, Punjab Province, Pakistan, 57000

**Abstract**

Alzheimer's disease (AD) is an irreversible devastative neurodegenerative disorder associated with progressive impairment of memory and cognitive functions. Its early diagnosis is crucial for the development of possible future treatment option(s). Structural magnetic resonance images (sMRI) plays an important role to help in understanding the anatomical changes related to AD especially in its early stages. Conventional methods require the expertise of domain experts and extract hand-picked features such as gray matter substructures and train a classifier to distinguish AD subjects from healthy subjects. Different from these methods, this paper proposes to construct multiple deep 2D convolutional neural networks (2D-CNNs) to learn the various features from local brain images which are combined to make the final classification for AD diagnosis. The whole brain image was passed through two transfer learning architectures; Inception version 3 and Xception; as well as custom Convolutional Neural Network (CNN) built with the help of separable convolutional layers which can automatically learn the generic features from imaging data for classification. Our study is conducted using cross-sectional T1-weighted structural MRI brain images from Open Access Series of Imaging Studies (OASIS) database to maintain the size and contrast over different MRI scans. Experimental results show that the transfer learning approaches exceed the performance of non-transfer learning based approaches demonstrating the effectiveness of these approaches for the binary AD classification task.

**Keywords**    Deep Learning, Transfer Learning, Artificial Neural Networks, Medical Imaging, Classification, MRI

## 1. Introduction

Cognitive decline associated with Alzheimer's disease (AD) is among the prevalent and widely studied phenomenon associated with the death of nerve cells, accumulation of amyloid plaques and neurofibrillary tangles, and loss of tissue throughout the brain that usually starts slowly and worsens over time. Seamless changes in the AD continuum take years if not decades to progress from normal cognition (NC) to mild cognitive impairment (MCI), with gradual evolution of clinically probable AD to confirmed AD [1-4]. Early and accurate detection and diagnosis of AD requires careful assessment by a medical domain expert, monitoring of patient history as well as physical and neurological examinations. The Mini–Mental State Examination (MMSE), which is a 30-point questionnaire [5–7], is a brief cognitive assessment tool commonly used to screen for dementia.

Neuroimaging modalities such as structural magnetic resonance imaging (sMRI), functional magnetic resonance imaging (fMRI) and positron emission tomography (PET) are used to provide the evidence that cognitive decline is structurally changing the brain [8-10]. They contain detailed information regarding the subcortical structures such as cerebral cortex, amygdala, hypothalamus, cerebellum; occipital, parietal and frontal lobes, corpus callosum, and thalamus. All these substructures of the brain are directly or indirectly involved in the distinctive cognitive and behavioral operations and a defect in the performance of them augments impaired decision making [11-15]. Specifically, it is known that changes including focal lesions and gray matter loss in the lobes can be understood using structural MRI scans [16].

Machine learning techniques have been widely used over the past decade for the analysis of neuroimaging data. Recently, a machine learning framework known as deep learning has received increased attention because of its ability in predicting various clinical outcomes of interest. Given the right infrastructure, deep learning algorithms are now arguably the best choice for classification of clinical phenotypes as they have the power to be a superior approach than traditional machine learning models for these tasks. Researchers have developed convolutional neural network (CNN) models which efficiently combine a number of approaches for successful pattern recognition and classification. CNN models have proven the ability to detect patterns within image data sets for common tasks such as classification with superior accuracy. Inside a CNN, a series of filters, with a size equivalent to a miniscule image patch, automatically search through the image to determine spatial patterns in the image. These filters can be learned and updated independently to detect critical information. Such approaches can be used in a straightforward fashion to train deep learning models on 2D and 3D MRI scans [17-33].

The datasets collected in neuroimaging studies are commonly very small, compared to the sizes of image classification datasets that are currently used to train neural networks for object recognition in 2D image analysis. To select discriminative features from a multitude of patches in each MR image still remains a challenging problem. Hence, in supervised learning setup, it is still incredibly important to be able to build network architectures capable of learning most discriminative features necessary for classification based on small datasets.

Motivated by the success of deep learning in computer vision and despite extensive research in the use of deep learning based methods for binary classification of AD using neuroimaging, there is still room for further exploration of these techniques especially for 2D classification architectures. In this study, we propose a 2D classification architecture based on the integration of several separable convolutional layers for cross-sectional sMRI brain image classification and disease diagnosis. Our architecture was designed and tested while addressing the paucity of data. The performance, in terms of binary classification, of the proposed architecture is compared with the existing transfer learning architectures Xception [35] and Inception version 3 [34] whose weights are pre-trained on Imagenet Large Scale Visual Recognition Challenge (ILSVRC) dataset. We assessed the performances of these architectures on a single modality which allows us to better apprehend its discriminating power. We use cross-sectional T1-weighted sMRI images from Open Access Series of Imaging Studies (OASIS) database for all the experiments [36].

The rest of this paper is organized as follows. In Section 2, we present the datasets in detail while the classification architectures are discussed in section 3. Experiments and results are presented in section 4 followed by discussion in section 5. Finally, a conclusion is given in section 6.

2. **Datasets**

We used OASIS-1 dataset of cross-sectional images to create subsets in all of our experiments. This set consists of a cross-sectional collection of 416 subjects aged 18 to 96. For each subject, 3 or 4 individual T1-weighted magnetization prepared rapid gradient-echo (MP-RAGE) images obtained in single scan sessions are included. The scans were acquired using a 1.5-T vision scanner in a single imaging session. The subjects are all right-handed and include both men and women. 100 of the included subjects over the age of 60 have been clinically diagnosed with very mild to moderate AD.

In this dataset, the diagnosis of AD is based on the clinical information that the subject has experienced gradual onset and progression of decline in memory and other cognitive and functional domains. Specifically, the clinical dementia rating (CDR) [37] is a dementia staging scale that rates subjects for impairment in each of the six domains: memory, orientation, judgment and problem solving, function in community affairs, home and hobbies, and personal care. A global CDR score is derived from individual ratings in each domain. A global CDR of 0 indicates no dementia, and CDRs of 0.5, 1, 2, and 3 represent very mild, mild, moderate, and severe dementia, respectively.

We use the atlas-registered gain field-corrected images in the dataset for all the experiments in this study. We define Non-Alzheimer's subject as one with CDR rating of zero, and Alzheimer's subject as one with CDR rating of 0.5, 1 or 2. Random samples of images of the two classes used for the experiments in this study are shown in figures Fig.1 and Fig. 2 below.

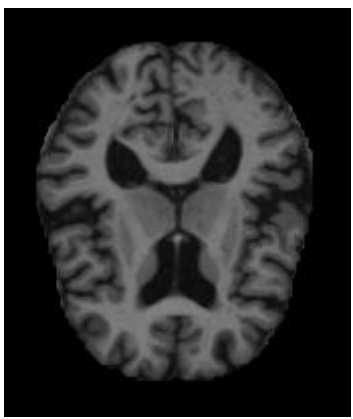     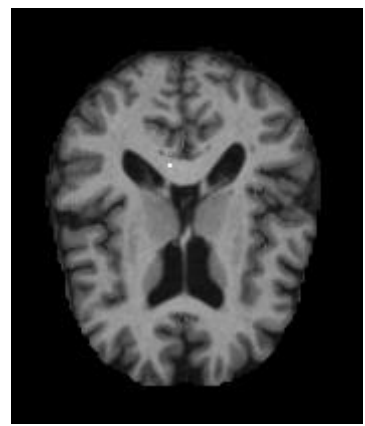

**Fig. 1** Alzheimer's Subject          **Fig. 2** Non-Alzheimer Subject

We used 96 slices for every subject selected visually from a 3D image after ignoring first and last 40 slices on the basis of the information they contain. Details of the datasets used for experiments in this study will be described in the next subsections.

## 2.1 Dataset-1

There are 90 subjects of both classes in this subset. CDR rating of Alzheimer's class instances is 0.5, 1 or 2. There are 60 subjects with CDR rating of 0.5, 28 subjects with CDR rating of 1, and 2 subjects with CDR rating of 2. The demographics of the subjects in this subset are listed in table-1.

**Table 1**     Demographics of subjects in Dataset-1

| No. of Subjects | | Age | | Gender | | MMSE | |
|---|---|---|---|---|---|---|---|
| Non-Alzheimer's Class | Alzheimer's Class | Non-Alzheimer's Class | Alzheimer's Class | Non-Alzheimer's Class | Alzheimer's Class | Non-Alzheimer's Class | Alzheimer's Class |
| 90 | 90 | Range: 62-94 Mean: 76.92 Standard Deviation: 8.55 | Range: 62-96 Mean: 76.93 Standard Deviation: 7.40 | Male: 24 Female: 66 | Male: 38 Female: 52 | Range: 25-30 Mean: 28.90 Standard Deviation: 1.24 | Range: 14-30 Mean: 24.05 Standard Deviation: 4.26 |

## 2.2 Dataset-2

We created this subset to study the class imbalance problem. There are 84 subjects of non-Alzheimer's class while 30 subjects of Alzheimer's class in this subset. The ratio between non-Alzheimer's and Alzheimer's subjects is 2.8:1 respectively. CDR rating of Alzheimer's class instances is 1 or 2. There are 28 subjects with CDR rating of 1 while 2 subjects with CDR rating of 2. The demographics of the subjects in this subset are listed in table-2.

**Table 2**  Demographics of subjects in Dataset-2

| No. of Subjects | | Age | | Gender | | MMSE | |
|---|---|---|---|---|---|---|---|
| Non-Alzheimer's Class | Alzheimer's Class | Non-Alzheimer's Class | Alzheimer's Class | Non-Alzheimer's Class | Alzheimer's Class | Non-Alzheimer's Class | Alzheimer's Class |
| 84 | 30 | Range: 65-94 Mean: 77.90 Standard Deviation: 7.98 | Range: 65-96 Mean: 78.03 Standard Deviation: 6.91 | Male: 22 Female: 62 | Male: 10 Female: 20 | Range: 25-30 Mean: 28.88 Standard Deviation: 1.22 | Range: 15-29 Mean: 21.23 Standard Deviation: 3.99 |

### 2.3 Dataset-3

We used the dataset from Marcia Hon et al [38]. As pointed out by Junhao Wen et al [17], there can be the problem of data leakage in this dataset which might have biased the performance upwards. The type of data leakage could potentially be an absence of an independent test set and a late split.

The authors [38] randomly picked 200 subjects, 100 each from the two classes. They used an entropy-based sorting mechanism to pick the most informative 32 slices from each 3D scan. That resulted in a total of 6400 images, 3200 for each class. 5,120 images are present in the training subset and 1,280 images are present in the validation subset.

### 3. Classification Architectures

CNNs are at the heart of modern deep learning revolution which has kickstarted after the 2012 ImageNet Large Scale Visual Recognition Challenge (ILSVRC) won by the Alexnet network.

The core of CNNs is the operation of convolution which learns local translationally invariant patterns in the input feature space. CNNs also take advantage of the fact that visual world is fundamentally spatially hierarchical by learning spatial hierarchies of patterns in their layers of which higher layers learn incrementally complex mappings of features learned by the lower layers.

Transfer learning [39] is a popular deep learning approach aimed at reusing a model developed for a task as a starting point for a model on another task. It is a popular way of dealing with problems with comparatively little data. In this work, we have used Xception and Inception version 3 neural network models as transfer learning models for the classification of subjects between the Alzheimer's and non-Alzheimer's groups. Figure Fig. 3 below shows the transfer learning model used in this study. We used sigmoid as an activation function to make the final decision based on the previous layers output.

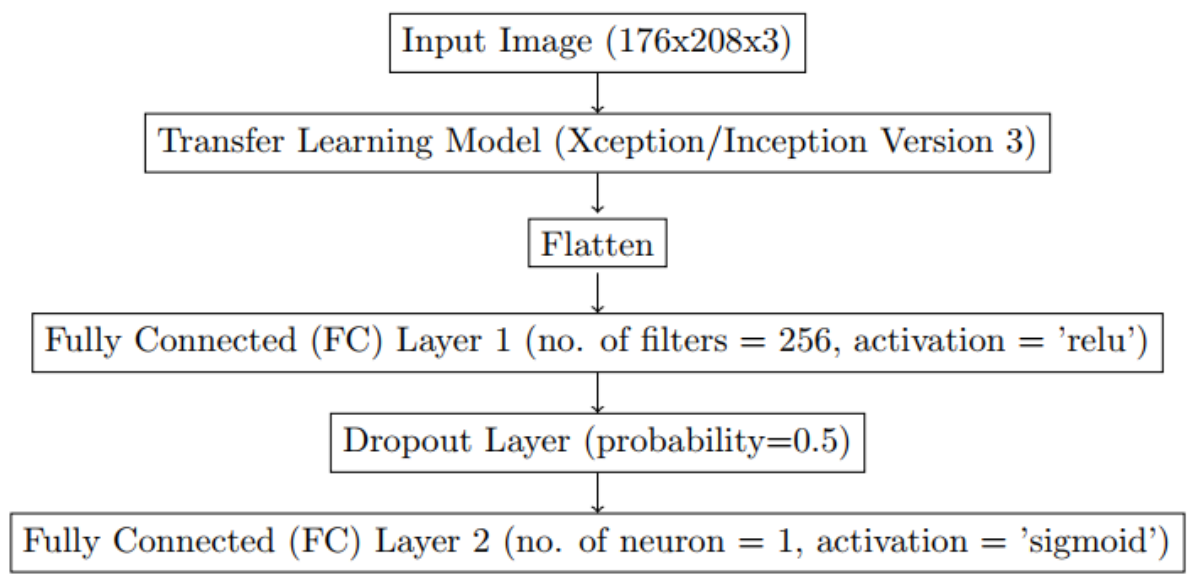

**Fig. 3** Transfer Learning Model

### 3.1 Inception Version 3 Model

In Inception modules, a convolutional layer attempts to learn filters to simultaneously and separately map cross-channel and spatial correlations. Inception version 3 uses the idea of replacing large size convolutional blocks with smaller sized ones, for example, a 5 x 5 convolution block is replaced by two 3 x 3 convolution blocks which results in the savings of computations. It also uses the notion of auxiliary classifiers, acting as regularizers, to combat the vanishing gradients problem by pushing the gradients to the lower layers to make them useful and improve the convergence of the network during training phase. Fig. 4 below shows a canonical Inception version 3 module.

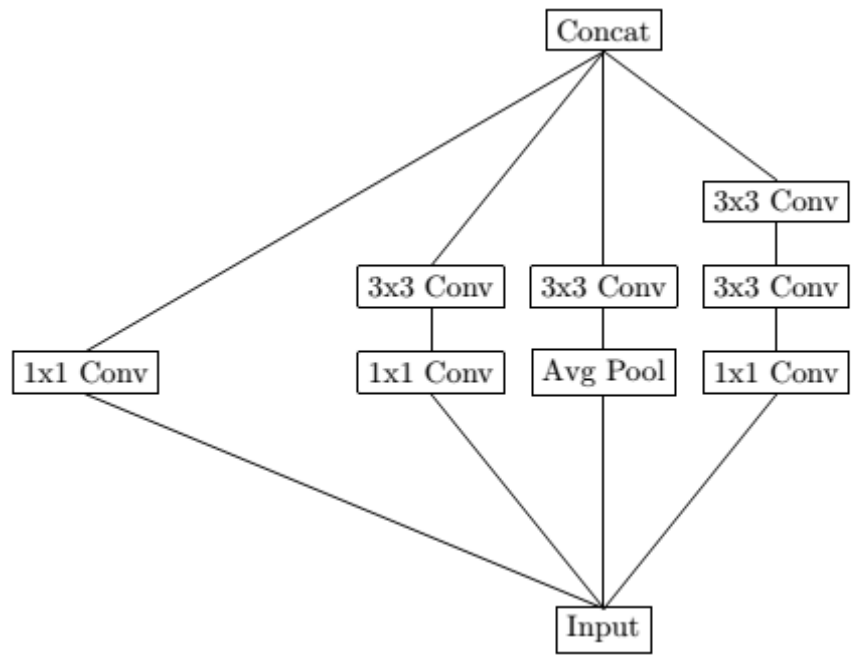

**Fig. 4** A canonical Inception version 3 module

### 3.2 Xception Model

Xception model is a novel deep learning neural network model which gets its inspiration from Inception. It uses the idea of depth-wise separable convolution which is a spatial convolution not using any non-linearity, independently performed, over channels of an input, followed by a pointwise convolution, hence projecting the channels output onto a new channel space.

### 3.3 Custom Model

We used a custom 2D-CNN model built with the help of depth-wise separable convolution, batch normalization, global average pooling, fully connected, and dropout layer(s) as shown in the figure Fig. 5 below. Our CNN architecture consists of nine depth-wise separable convolutional and batch-normalization layers. Rectified linear units (ReLU) were used as activation functions for the convolutional layers. There are two fully-connected layers. The activation function for the second fully connected layer with a single neuron is sigmoid which will classify the target into two categories using a threshold of 0.5. Batch normalization is a regularization technique that normalizes the input of a given set of layers (the normalization is done using the mean and standard deviation of a batch). This procedure acts as a regularizer and in some cases eliminates the need for the dropout. It helps fight the exploding gradient phenomenon and allows using much higher learning rates. To some extent, it frees the model from initialization problems. Dropout randomly and independently drops neurons, setting their output value to zero along with their connections. It aims to make the network less complex and thus less prone to overfitting.

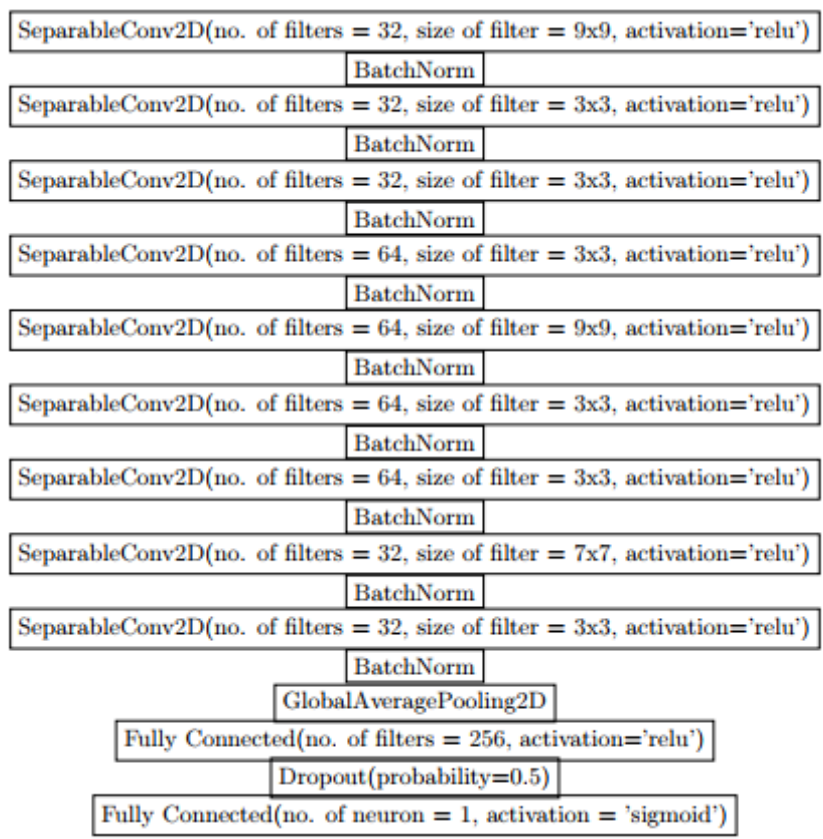

**Fig. 5** Custom CNN architecture

4. **Experiments and Results**

We performed our experiments on dataset-1, dataset-2 and dataset-3. We use 2, 5 and 9-fold cross-validation procedures to get better approximation of prediction performances on dataset-1. We use 6-fold cross-validation to study the class imbalance problem on dataset-2. Dataset-3 was splitted for 5-folds and studied in the context of the data leakage problem in the neuroimaging studies.

When working on datasets 1 and 2, as a preprocessing step, we applied width shift, height shift and horizontal flipping as well as scale normalization to augment the dataset for training and only scale normalization on the validation set.

When working on dataset 3, we applied scale normalization as a preprocessing step for both training and validation sets.

For the evaluation of results on dataset-1, we computed the following performance metrics: Accuracy, Sensitivity, Specificity, Precision, $F_1$-score and Matthews correlation coefficient (MCC) defined as under.

$$\text{Accuracy} = \frac{TP+TN}{TP+TN+FP+FN}$$

$$\text{Sensitivity} = \frac{TP}{TP+FN}$$

$$\text{Specificity} = \frac{TN}{TN+FP}$$

$$\text{Precision} = \frac{TP}{TP+FP}$$

$$F_1\text{-score} = \frac{2TP}{2TP+FP+FN}$$

$$\text{MCC} = \frac{TPxTN - FPxFN}{\sqrt{(TP+FP)(TP+FN)(TN+FP)(TN+FN)}}$$

Here accuracy is the proportion of correctly classified subjects among the whole population. Sensitivity also called recall is the proportion of Alzheimer's class subjects correctly classified among the total number of Alzheimer's class subjects. Specificity is the proportion of non-Alzheimer's class subjects correctly classified among the total number of non-Alzheimer's class subjects. $F_1$-score is the harmonic mean of precision and sensitivity. The ideal value of these metrics is 1 and is the target for the models in this study. We define true positive (TP) as hit that the model predicted Alzheimer's subject class correctly. False positive (FP) also called Type-I error happens when model incorrectly predicts a non-Alzheimer's class instance to be an instance of Alzheimer's class. False negative (FN) also called Type-II error happens when model incorrectly predicts an instance of Alzheimer's class to be an instance of non-Alzheimer's class. Finally, true negative (TN) occurs when the model predicts an instance of a non-Alzheimer's class correctly. In general, we use the models with the best validation set accuracies to compute these metrics.

We used the best validation set accuracy as the metric for comparing the architectures on the dataset-2 and dataset-3.

We will now present the details of all experiments accompanied with the results in the following subsections.

### 4.1 Dataset-1 Two-fold Cross-Validation Procedure

For this experiment, we split the dataset into two folds. There are 45 subjects in the training and validation subsets. Each class contains 8640 samples so there are 4320 samples of each class in both training and validation subsets. The learning rate for this experiment is set to $10^{-4}$. We use binary cross-entropy as a loss function and stochastic gradient descent (SGD) as the optimization algorithm. We set the initial learning rate of SGD to $10^{-4}$ and employed step decay to decrease the learning rate after every epoch. We use gradient clipping to clip the values of the gradients above 0.5 and set the clipping norm to a value of 1. We train each model for 30 epochs and set the batch size to 16. The results of the three models with mean and standard deviation values are given in the tables Table-4, Table-5 and Table-6.

**Table 4**     Results of Inception Version 3 architecture on dataset-1 for 2-fold cross-validation procedure

| Accuracy | Sensitivity (Recall) | Specificity | Precision | $F_1$-Score | MCC |
|---|---|---|---|---|---|
| 0.6298±0.0122 | 0.6305±0.0132 | 0.6294±0.0107 | 0.6295±0.0120 | 0.6301±0.0124 | 0.2596±0.0243 |

**Table 5**     Results of Xception architecture on dataset-1 for 2-fold cross-validation procedure

| Accuracy | Sensitivity (Recall) | Specificity | Precision | $F_1$-Score | MCC |
|---|---|---|---|---|---|
| 0.6384±0.0054 | 0.6389±0.0054 | 0.6380±0.0053 | 0.6383±0.0054 | 0.6385±0.0054 | 0.2768±0.0107 |

**Table 6**     Results of Custom CNN architecture on dataset-1 for 2-fold cross-validation procedure

| Accuracy | Sensitivity (Recall) | Specificity | Precision | F$_1$-Score | MCC |
|---|---|---|---|---|---|
| 0.6082±0.0021 | 0.6082±0.0121 | 0.6082±0.0162 | 0.6083±0.0051 | 0.6082±0.0035 | 0.2165±0.0041 |

### 4.2 Dataset-1 Five-Fold Cross-Validation Procedure

For this experiment, we split the dataset into five folds. There are 72 subjects in the training and 18 subjects in the validation subset. The learning rate for this experiment is set to $10^{-4}$. We use binary cross-entropy as a loss function and stochastic gradient descent (SGD) as the optimization algorithm. We set the initial learning rate of SGD to $10^{-4}$ and employed step decay to decrease the learning rate after every epoch. We use gradient clipping to clip the values of the gradients above 0.5 and set the clipping norm to a value of 1. We train each model for 30 epochs and set the batch size to 16. The results of the three models with mean and standard deviation values are given in the tables Table-7, Table-8 and Table-9.

**Table 7** Results of Inception Version 3 architecture on dataset-1 for 5-fold cross-validation procedure

| Accuracy | Sensitivity (Recall) | Specificity | Precision | F$_1$-Score | MCC |
|---|---|---|---|---|---|
| 0.6454±0.0342 | 0.6461±0.0343 | 0.6447±0.0341 | 0.6452±0.0341 | 0.6456±0.0342 | 0.2908±0.0684 |

**Table 8** Results of Xception architecture on dataset-1 for 5-fold cross-validation procedure

| Accuracy | Sensitivity (Recall) | Specificity | Precision | F$_1$-Score | MCC |
|---|---|---|---|---|---|
| 0.6453±0.0339 | 0.6478±0.0352 | 0.6428±0.0331 | 0.6445±0.0336 | 0.6461±0.0343 | 0.2906±0.0678 |

**Table 9** Results of Custom CNN architecture on dataset-1 for 5-fold cross-validation procedure

| Accuracy | Sensitivity (Recall) | Specificity | Precision | $F_1$-Score | MCC |
|---|---|---|---|---|---|
| 0.6219±0.0400 | 0.6239±0.0402 | 0.6199±0.0428 | 0.6216±0.0411 | 0.6227±0.0401 | 0.2439±0.0801 |

### 4.3 Dataset-1 Nine-Fold Cross-Validation Procedure

For this experiment, we split the dataset into nine folds. There are 80 subjects in the training and 10 subjects in the validation subset. The learning rate for this experiment is set to $10^{-4}$. We use binary cross-entropy as a loss function and stochastic gradient descent (SGD) as the optimization algorithm. We set the initial learning rate of SGD to $10^{-4}$ and employed step decay to decrease the learning rate after every epoch. We use gradient clipping to clip the values of the gradients above 0.5 and set the clipping norm to a value of 1. We train each model for 30 epochs and set the batch size to 16. The results of the three models with mean and standard deviation values are given in the tables Table-10, Table-11 and Table-12.

**Table 10**    Results of Inception Version 3 architecture on dataset-1 for 9-fold cross-validation procedure

| Accuracy | Sensitivity (Recall) | Specificity | Precision | $F_1$-Score | MCC |
|---|---|---|---|---|---|
| 0.6406±0.0340 | 0.6413±0.0338 | 0.6399±0.0342 | 0.6404±0.0340 | 0.6408±0.0339 | 0.2812±0.0679 |

**Table 11**    Results of Xception architecture on dataset-1 for 9-fold cross-validation procedure

| Accuracy | Sensitivity (Recall) | Specificity | Precision | $F_1$-Score | MCC |
|---|---|---|---|---|---|
| 0.6487±0.0365 | 0.6488±0.0369 | 0.6486±0.0362 | 0.6487±0.0364 | 0.6488±0.0367 | 0.2975±0.0731 |

**Table 12**    Results of Custom CNN architecture on dataset-1 for 9-fold cross-validation procedure

| Accuracy | Sensitivity (Recall) | Specificity | Precision | F$_1$-Score | MCC |
|---|---|---|---|---|---|
| 0.6479±0.0447 | 0.6492±0.0384 | 0.6466±0.0516 | 0.6481±0.0466 | 0.6486±0.0424 | 0.2958±0.0895 |

### 4.4 Dataset-2 Six-Fold Cross-Validation Procedure

For this experiment, we split the dataset into six folds. There are 95 subjects in the training and 19 subjects in the validation subset. There are 70 subjects of non-Alzheimer's class and 25 subjects of Alzheimer's class in the training subset while 14 subjects of non-Alzheimer's class and 5 subjects of Alzheimer's class in the validation subset. The learning rate for this experiment is set to $10^{-4}$. We use binary cross-entropy as a loss function and stochastic gradient descent (SGD) as the optimization algorithm. We set the initial learning rate of SGD to $10^{-4}$ and employed step decay to decrease the learning rate after every epoch. We use gradient clipping to clip the values of the gradients above 0.5 and set the clipping norm to a value of 1. We train each model for 30 epochs and set the batch size to 16. The best validation set accuracies for the three models in each fold are given in the tables Table-13, Table-14 and Table-15. Figures Fig. 6, Fig. 7, Fig. 8, Fig. 9, Fig. 10 and Fig. 11 shows random training and validation sets accuracies and loss plots for the three models.

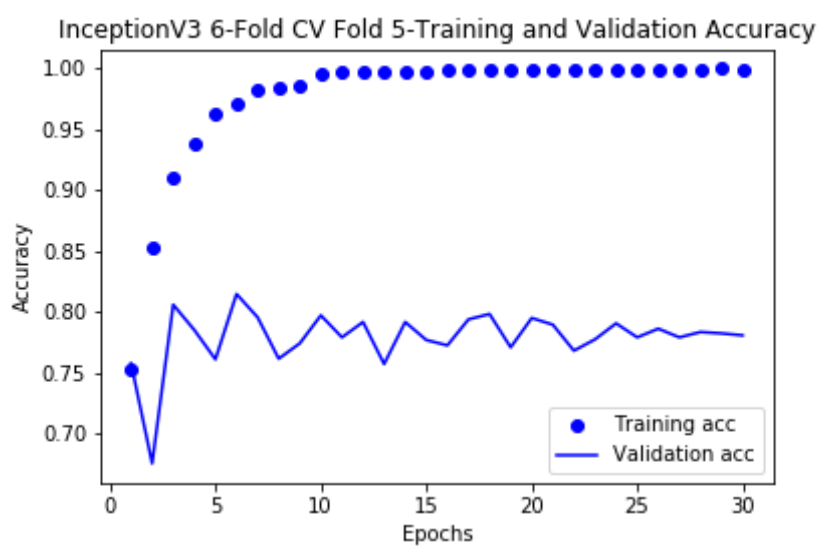

**Fig. 6** Accuracy plot for the Inception version 3 architecture on dataset-2

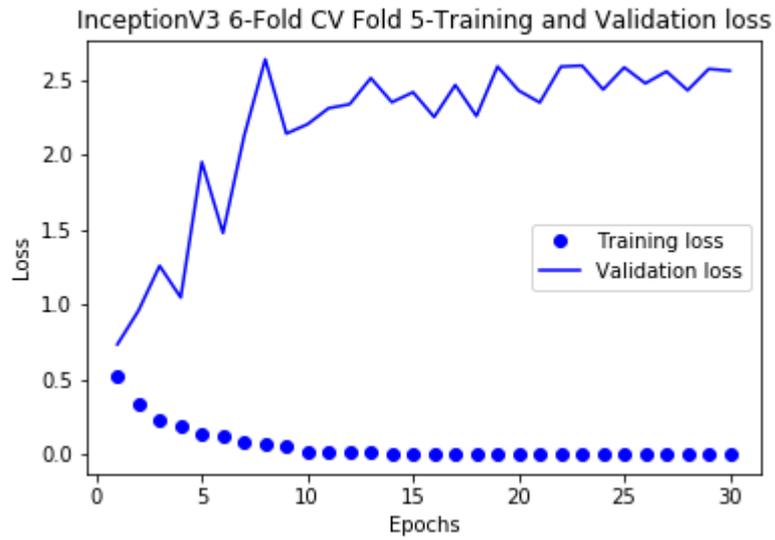

**Fig. 7** Loss plot for the Inception version 3 architecture on dataset-2

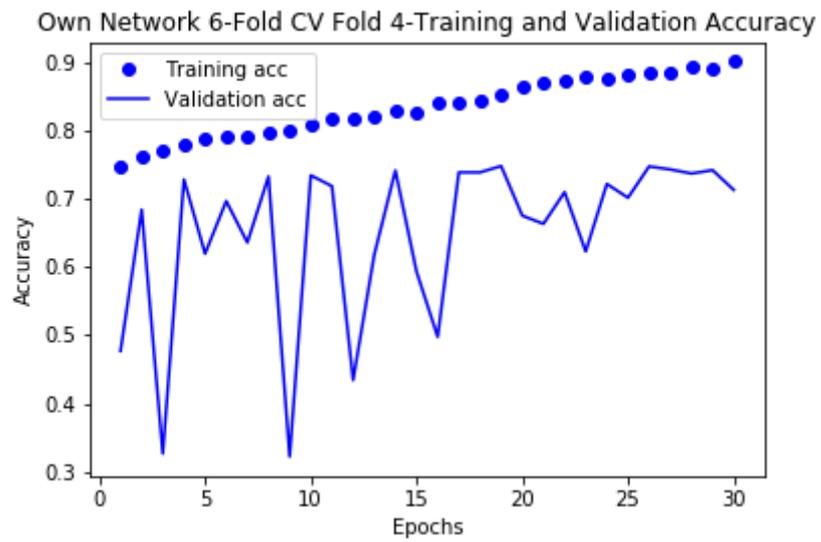

**Fig. 8** Accuracy plot for the Custom CNN architecture on dataset-2

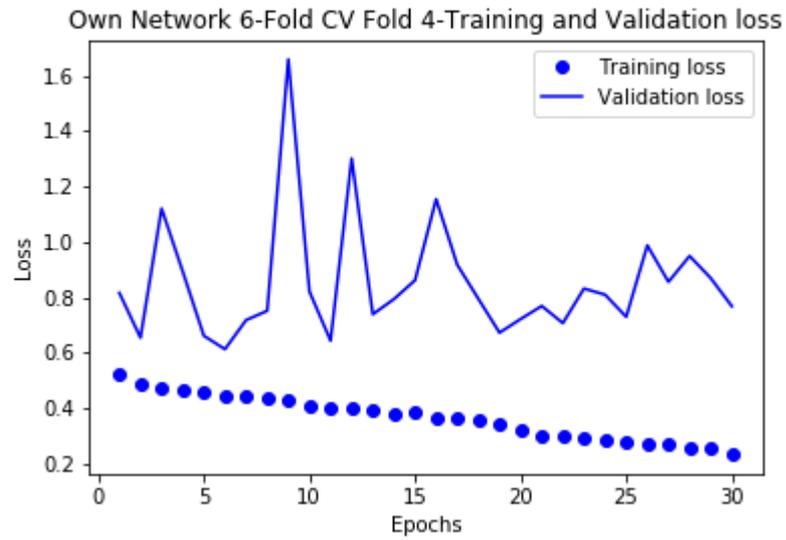

**Fig. 9** Loss plot for the Custom CNN architecture on dataset-2

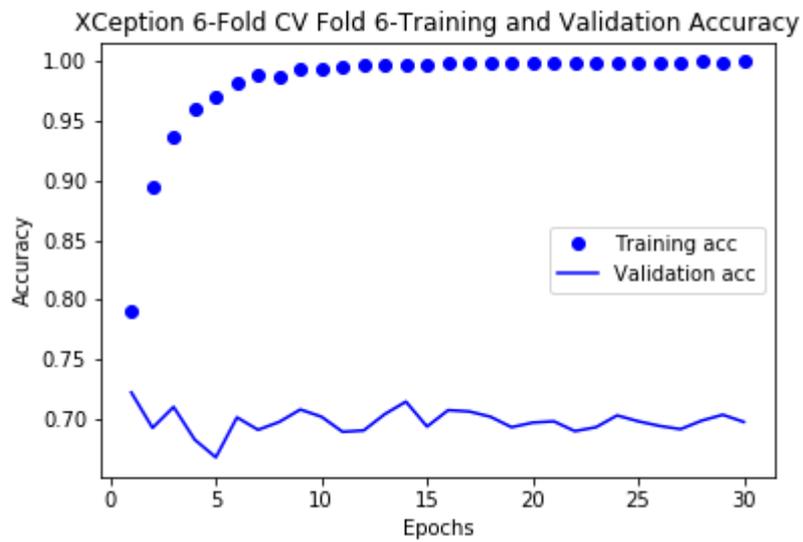

**Fig. 10**    Accuracy plot for the Xception architecture on dataset-2

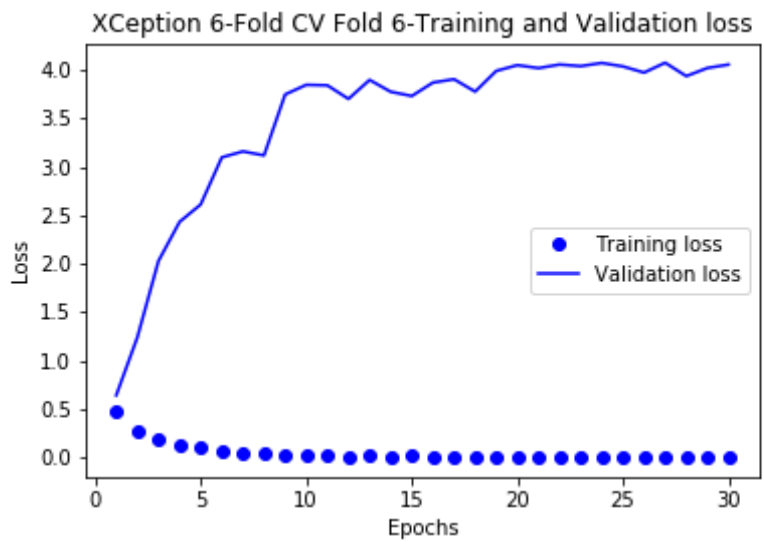

**Fig. 11**     Loss plot for the Xception architecture on dataset-2

**Table 13**     Results of Inception Version 3 architecture on dataset-2 for 6-fold cross-validation procedure

| Fold-1 | Fold-2 | Fold-3 | Fold-4 | Fold-5 | Fold-6 |
|---|---|---|---|---|---|
| 74.45% | 77.30% | 77.63% | 77.25% | 81.14% | 74.01% |

**Table 14**     Results of Xception architecture on dataset-2 for 6-fold cross-validation procedure

| Fold-1 | Fold-2 | Fold-3 | Fold-4 | Fold-5 | Fold-6 |
|---|---|---|---|---|---|
| 72.97% | 77.25% | 80.26% | 79.61% | 81.80% | 72.26% |

**Table 15**     Results of Custom CNN architecture on dataset-2 for 6-fold cross-validation procedure

| Fold-1 | Fold-2 | Fold-3 | Fold-4 | Fold-5 | Fold-6 |
|---|---|---|---|---|---|
| 75.11% | 76.75% | 79.00% | 74.89% | 82.79% | 74.45% |

### 4.5 Dataset-3 Five-Fold Cross-Validation Procedure

For this experiment, the dataset was splitted by the authors [38] into five folds. There are 5120 2-D slices in the training subset and 1280 2-D slices in the validation subset.

We use binary cross-entropy as a loss function and stochastic gradient descent (SGD) as the optimization algorithm. We set the initial learning rate of SGD to $10^{-4}$ and employed step decay to decrease the learning rate after every epoch. We use gradient clipping to clip the values of the gradients above 0.5 and set the clipping norm to a value of 1. We trained the Xception and Inception version 3 models for 100 epochs and set the batch size to 8. We trained the custom CNN architecture for 60 epochs and set the batch size to 16. The best validation set accuracies for the three models in each fold are given in the tables Table-16, Table-17 and Table-18. Figures Fig. 12, Fig. 13, Fig. 14, Fig. 15, Fig. 16 and Fig. 17 shows random training and validation sets accuracies and loss plots for the three models.

**Table 16**   Results of Inception Version 3 architecture on dataset-3 for 5-fold cross-validation procedure

| Fold-1 | Fold-2 | Fold-3 | Fold-4 | Fold-5 |
|---|---|---|---|---|
| 99.30% | 98.91% | 99.06% | 98.91% | 99.45% |

**Table 17**   Results of Xception architecture on dataset-3 for 5-fold cross-validation procedure

| Fold-1 | Fold-2 | Fold-3 | Fold-4 | Fold-5 |
|---|---|---|---|---|
| 97.81% | 97.97% | 97.66% | 98.44% | 97.97% |

**Table 18**   Results of Custom CNN architecture on dataset-3 for 5-fold cross-validation procedure

| Fold-1 | Fold-2 | Fold-3 | Fold-4 | Fold-5 |
|---|---|---|---|---|
| 95.55% | 95.00% | 95.39% | 93.59% | 93.67% |

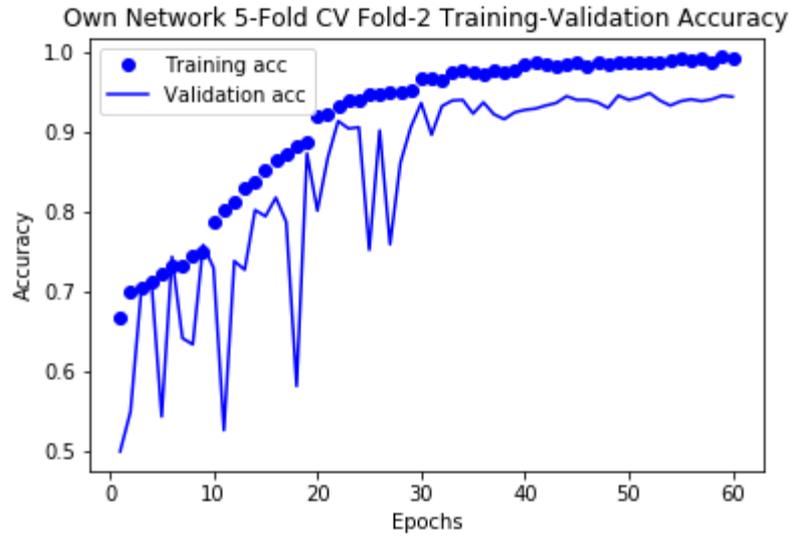

**Fig. 12**     Accuracy plot for the Custom CNN architecture on dataset-3

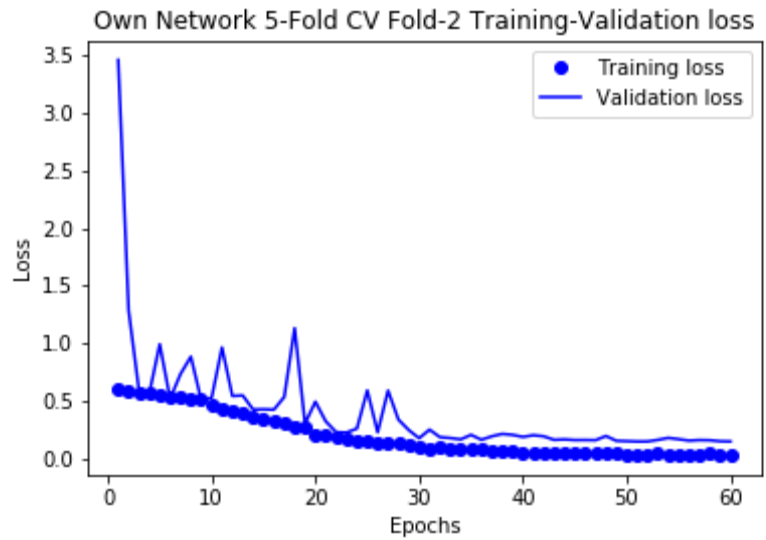

**Fig. 13**     Loss plot for the Custom CNN architecture on dataset-3

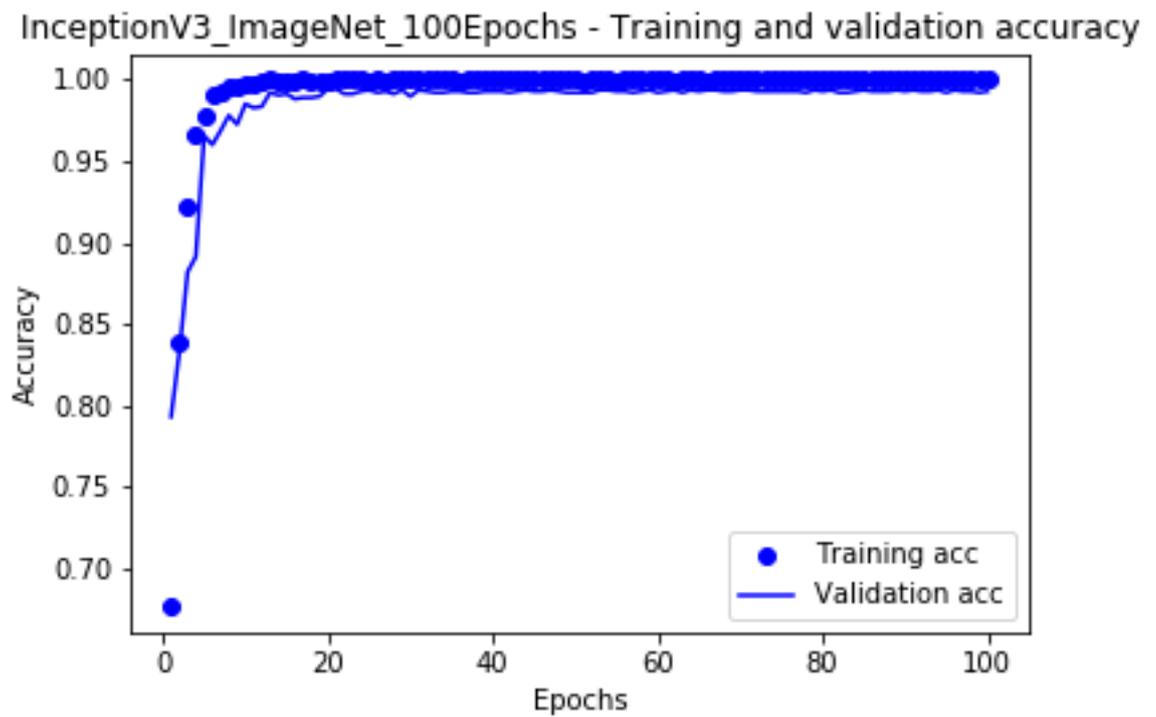

**Fig. 14**     Accuracy plot for the Inception version 3 architecture on dataset-3

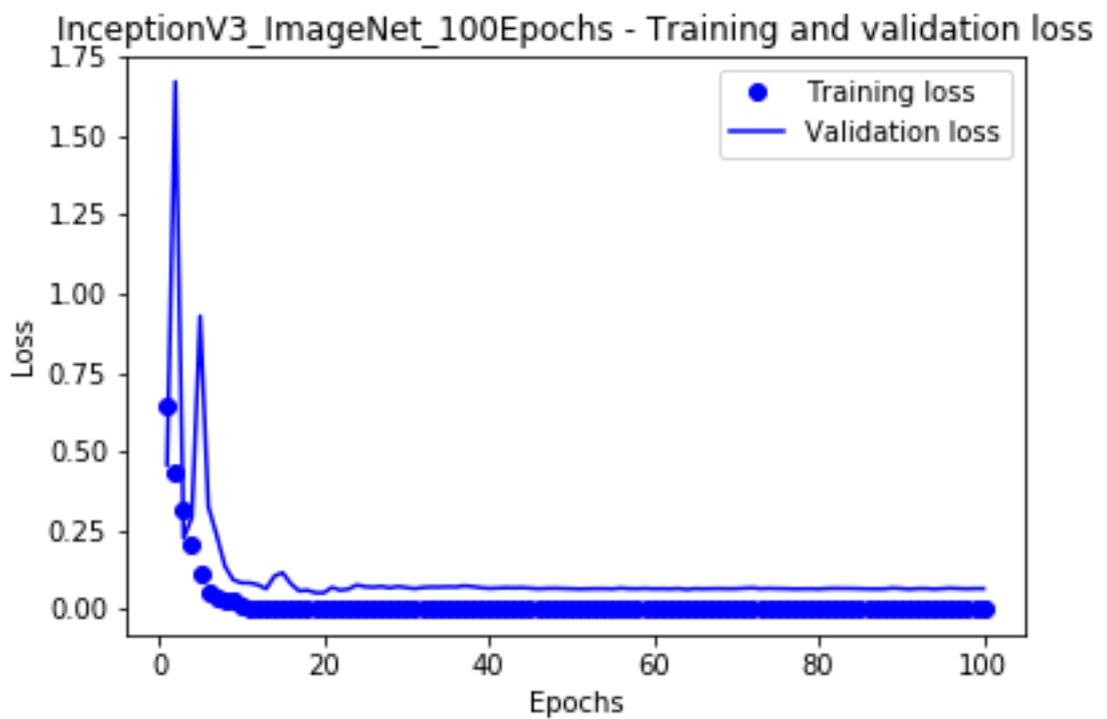

**Fig. 15**     Loss plot for the Inception version 3 architecture on dataset-3

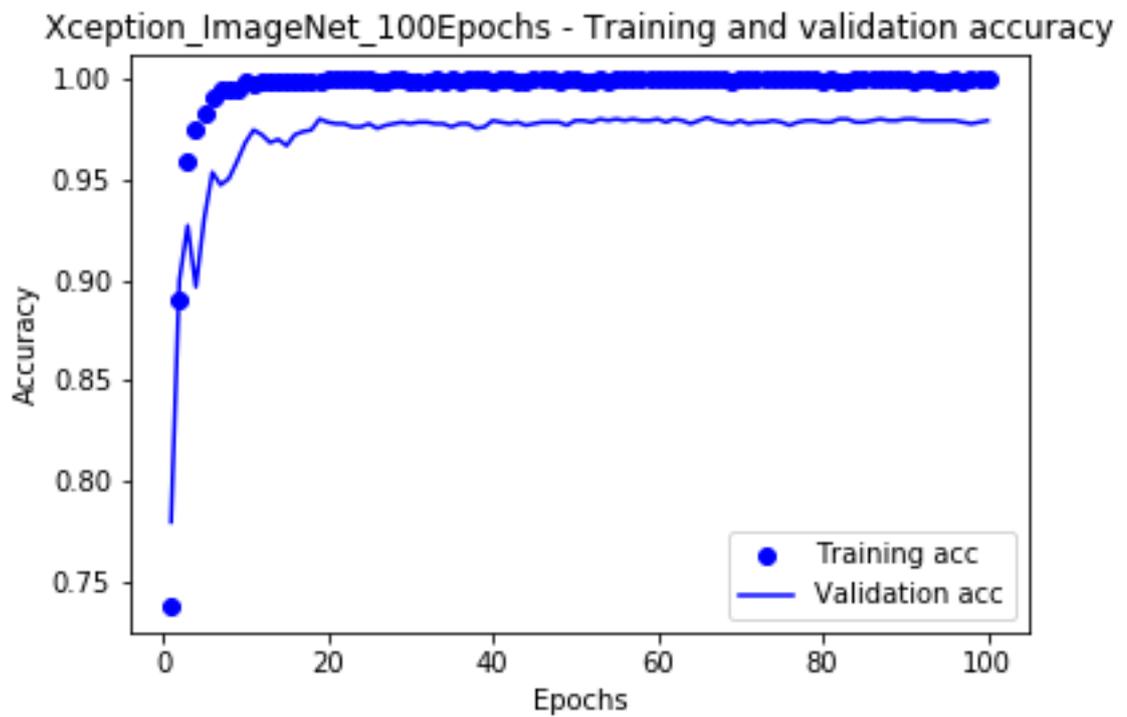

**Fig. 16**     Accuracy plot for the Xception architecture on dataset-3

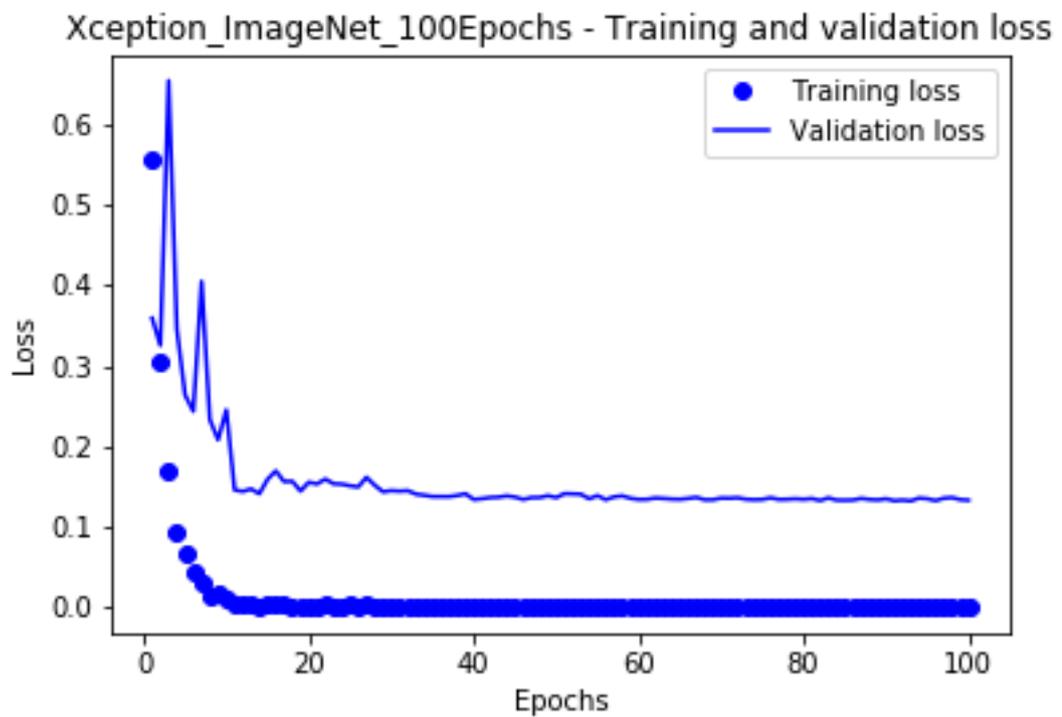

**Fig. 17**     Loss plot for the Xception architecture on dataset-3

5. **Discussion**

The present studies contains two main contributions. First, we compared the performance differences among transfer learning and non-transfer learning based approaches in deep neural networks. Second, we study the data imbalance and data leakage problems which are present in the arena of neuroimaging based studies in AD.

We have chosen the T1-weighted MRI images for our experiments due to their versatility and ease of availability. Apart from T1-weighted MRI imaging modality, Diffusion tensor imaging (DTI) is a recent MRI modality based on the motion of water molecules by quantifying the random water diffusion in cerebral gray and white matters. Quantifying the neurodegeneration in the hippocampus area leads to measure the pathological atrophy change, and discriminate AD subjects from those with MCI and NC. MCI is the prodromal stage of AD, landmarks with group differences between AD and NC subjects are the potential atrophy locations in brain MR images of MCI subjects. Shrinkage of hippocampal region which accompanies the development of AD is observable on structural MRI and DTI modalities. In essence, hippocampal region is one of the biomarkers of AD. It is worth mentioning that the early stage of AD would induce miniscule structural changes in local brain regions, instead of the whole brain [26]. Hence, feature representations defined at Region of Interest (ROI)-level or whole-image-level may not be effective in characterizing the early AD-related structural changes of the brain [28]. In addition, alterations in AD are not confined to the hippocampus and extend to other regions in the temporal, parietal and frontal lobes. The progress of the decline can happen in the brain including frontotemporal lobar degeneration associated with language impairment and posterior cortical atrophy relevant to visual impairment [30].

We realize that the most difficult classification tasks are mild AD vs. severe AD, and mild AD vs. NC [19]. We further verify that to be able to have a classifier that generalizes well enough it has to be trained on a large enough dataset. This is especially true for the CNN classifiers [18]. It is known that the abnormal brain regions relevant to AD might not well fit to the pre-defined ROIs. The 3D voxel-wise features can alleviate this problem, but given their huge dimensionality, far more features than training subjects, they may lead to low classification performances. The cortical thickness and hippocampus shape features neglect the correlated variations of the whole brain structure affected by AD. In addition, the correctness of extracted features highly depends on image preprocessing steps requiring the domain knowledge of an expert [21]. We further learned that a successful classification model has to learn both the local information of image patches and the global information of multiple landmarks by assigning the class labels at the subject-level. Since the structural changes in the brain induced by AD in its early stages could be subtle and distribute in different areas of the brain, we attempt to learn local to global feature representations for the experiments.

The raw brain images contain a lot of information and noise to be directly used for the task of classification. Thus, it is necessary to extract representative features for image classification. However, how to select discriminative patches from a large number of patches in each MR image still remains a problem of grand scale and proportions. Most of the existing representations are based on engineered and empirically predefined features independent of subsequent classifier learning procedure. Due to the possible heterogeneous nature of features and classifiers, the pre-defined features may lead to sub-optimal learning performance for brain disease diagnosis. Training a classifier independent from the feature extraction process may lead to sub-optimal learning performance, due to the possible heterogeneous nature of classifier and features. In this context, CNNs have become very popular for automatically learning representations from large collections of static images.

From the experiments, we learned that CNNs cannot be successfully trained without intensity rescaling. In addition, we applied augmentation to improve the classification performances. We learned that data imbalance is a critical issue and provides biased performances in favor of a certain class especially after performing the experiments on the dataset-2 as we got higher validation accuracies on the non-Alzheimer's class in comparison with the Alzheimer's class instances. The performance of transfer learning architectures is found to be better on all the metrics on the dataset-1 when compared to the custom CNN architecture. However, the losses of training and validating the transfer learning architectures are higher, in general, in comparison with the custom CNN architecture which can be explained by the overfitting problem. We further found the biased performances of architectures on dataset-3. As mentioned by Junhao Wen et al [17], there can be the problem of data leakage in this case. Due to problems in understanding this phenomenon, the creators of the dataset-3 Marcia Hon et al [38] have probably mixed the samples of a subject in both training and validation sets.

It is not easy to visualize the learned features by the proposed method for interpretation of the brain regions relevant to neurodegenerative disease (i.e., AD or MCI) for the clinical application. The learned features have no sufficient clinical information to find the related ROIs for clinical understanding of brain abnormalities. However, we plotted some of the patterns learned by the proposed custom CNN model as shown in the figures Fig. 18, Fig. 19, Fig. 20, Fig. 21, Fig. 22, Fig. 23, Fig. 24 and Fig. 25.

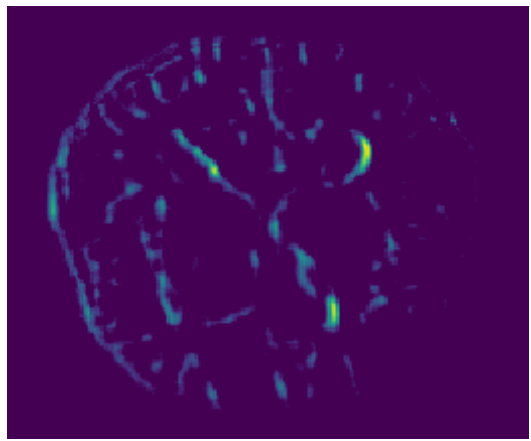

**Fig. 18**  Eighth channel of the activation of the first layer of the proposed custom CNN model on the AD instance

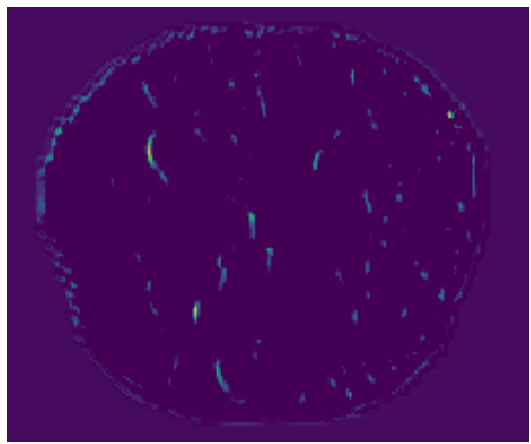

**Fig. 19**  Fifth channel of the activation of the fifth layer of the proposed custom CNN model on the AD instance

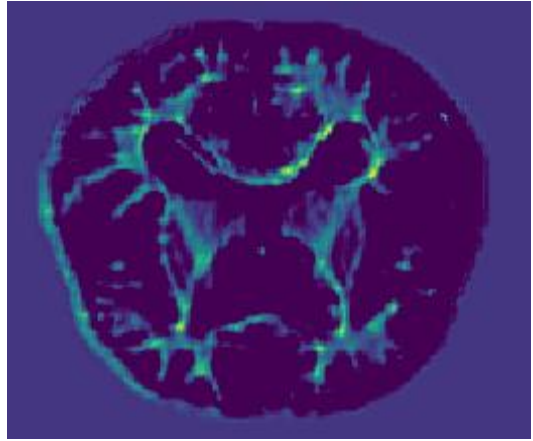

**Fig. 20**     First channel of the activation of the fifth layer of the proposed custom CNN model on the AD instance

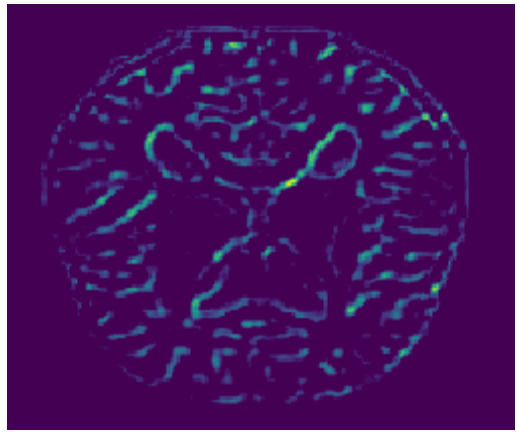

**Fig. 21**     First channel of the activation of the first layer of the proposed custom CNN model on the AD instance

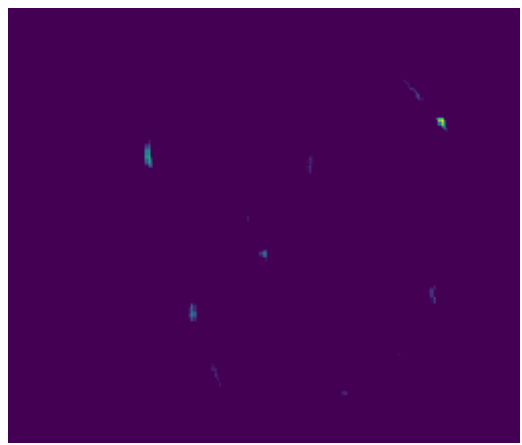

**Fig. 22**     Ninth channel of the activation of the first layer of the proposed custom CNN model on the AD instance

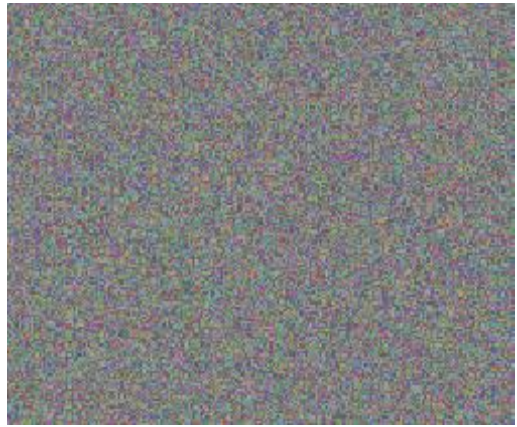

**Fig. 23**    Pattern that the seventh channel in layer separable_conv2D_16 of the proposed custom CNN model responds to maximally

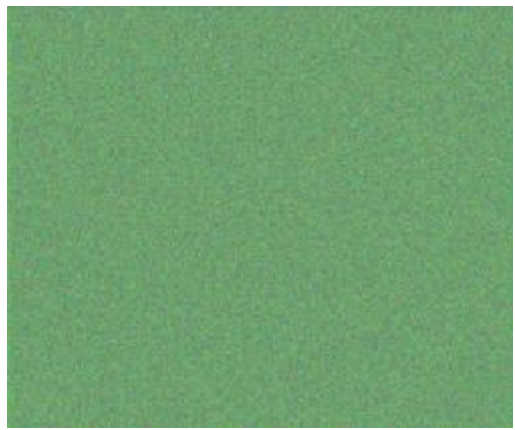

**Fig. 24**    Pattern that the zeroth channel in layer separable_conv2D_10 of the proposed custom CNN model responds to maximally

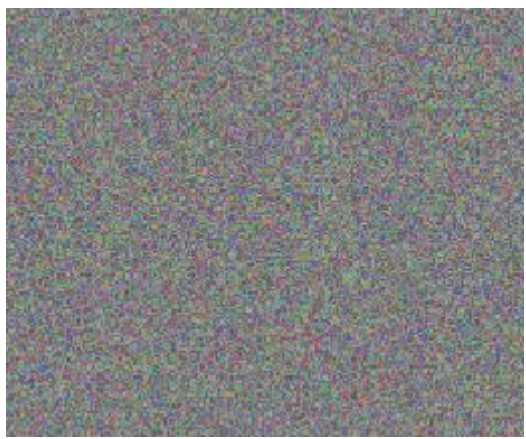

**Fig. 25**    Pattern that the zeroth channel in layer separable_conv2D_16 of the proposed custom CNN model responds to maximally

## 6. Conclusion

In this work, we compared and contrasted the performance of deep learning architectures on the AD binary classification problem using T1-weighted sMRI images. We proposed a custom CNN model built with separable convolutional layers and compared its performance on three datasets with transfer learning architectures Inception version 3 and Xception that are pretrained using the imagenet dataset. We found the performance of transfer learning architectures on this task to be better. We further studied the class imbalance and data leakage problems and found that they tend towards increasing the classification performance bias.

This work can be extended in a number of ways. As a future work, we will explore other datasets and imaging modalities such as functional MRI, DTI etc and multiclass classification problem of AD comprising of healthy control subjects, mild cognitive impairment (MCI) subjects and AD subjects with higher CDR rating. We will also explore more holistic approaches to this problem such as exploring new network architectures especially 3D architectures as well as getting insights to the learning of neural network architectures through different visualization approaches.